\documentclass{article}

\usepackage{PRIMEarxiv}

\usepackage[utf8]{inputenc} 
\usepackage[T1]{fontenc}    
\usepackage{hyperref}       
\usepackage{url}            
\usepackage{booktabs}       
\usepackage{amsfonts}       
\usepackage{nicefrac}       
\usepackage{microtype}      
\usepackage{lipsum}
\usepackage{fancyhdr}       
\usepackage{graphicx}       
\usepackage{amsmath}
\graphicspath{{media/}}     

\title{Deep generative models as the probability transformation functions
\thanks {This preprint has not undergone peer review or any post-submission improvements or corrections. The Version of Record of this contribution will be published in "ICIST 2025 Springer Proceedings"} }

\author{
    Vitalii Bondar \\
    Cherkasy State Technological University \\
    \texttt{v.v.bondar.asp24@chdtu.edu.ua} \\
 \And
    Vira Babenko \\
    Cherkasy State Technological University \\
    \texttt{v.babenko@chdtu.edu.ua} \\
 \And
    Roman Trembovetskyi \\
    Cherkasy State Technological University \\
    \texttt{roman.tremb@gmail.com} \\
 \And
    Yurii Korobeinyk \\
    Cherkasy State Technological University \\
    \texttt{yu.o.korobeinyk.asp24@chdtu.edu.ua} \\
 \And
    Viktoriya Dzyuba \\
    Cherkasy Bohdan Khmelnytsky National University \\
    \texttt{viktoriya.dzyuba15@gmail.com} \\
}

\begin{document}
\maketitle

\begin{abstract}
This paper introduces a unified theoretical perspective that views deep generative models as probability transformation functions. Despite the apparent differences in architecture and training methodologies among various types of generative models –- autoencoders, autoregressive models, generative adversarial networks, normalizing flows, diffusion models, and flow matching -- we demonstrate that they all fundamentally operate by transforming simple predefined distributions into complex target data distributions. This unifying perspective facilitates the transfer of methodological improvements between model architectures and provides a foundation for developing universal theoretical approaches, potentially leading to more efficient and effective generative modeling techniques.
\end{abstract}


\section{Introduction}
Deep generative models have gained significant prominence in recent years. The widespread adoption of these models for producing realistic images, audio, and video content, coupled with the emergence of large language models, has sparked considerable research interest. These advancements have been facilitated by increased computational capabilities and the accessibility of extensive datasets, enabling the implementation of generative models across diverse domains \cite{bond2021deep, yang2024_diff}.

Generative models have emerged as essential tools across diverse domains due to their capacity to synthesize novel data. In the medical field, these models facilitate protein structure prediction, enhance diagnostic precision through synthetic imaging, and accelerate drug discovery processes. Within engineering applications, they enable the development of innovative materials with precise specifications, particularly benefiting the aerospace and automotive sectors. Furthermore, these models have transformed creative industries by enabling the generation of photorealistic character designs, musical compositions, and visual effects \cite{gozalo2023survey, lipman2022flow}.

Generative models, while addressing the common task of data generation, employ diverse approaches to approximate data distributions. Although several classification attempts have been proposed in the literature \cite{bond2021deep, goodfellow2016nips}, a standardized taxonomy of generative models based on their characteristics remains undefined.

The primary categorization of generative models distinguishes between explicit and implicit density distribution representations. Models with explicit density distribution representation offer dual functionality: they can generate instances from the learned distribution and evaluate the probability of specific instances belonging to that distribution. Normalizing flows serve as a prominent example of this category. In contrast, models with implicit density distribution representation, such as generative adversarial networks, encode distributional knowledge within their parameters, making it inaccessible for direct probability estimation.

This paper examines the most prevalent classes of deep generative models: autoencoders, autoregressive models, generative adversarial networks, normalizing flows, diffusion models, and flow matching.

The field of generative models currently exists as disconnected domains of knowledge, with various types of generative models developing independently and rarely intersecting in their methodological approaches. Despite their apparent differences, we argue that most generative models can be unified under a common theoretical framework. This unification would facilitate the transfer of scientific advances between different types of generative models.

Our main contribution is the introduction of a unified perspective that views generative models as probability transformation functions. In this framework, each model operates as a specific function that maps a simple distribution to the target data distribution, while the characteristics of particular model types are defined by their optimization approaches.

This research presents a comprehensive framework that bridges various types of machine learning generative models, establishing a common theoretical foundation for their analysis and development.

\section{Background}

\subsection{Autoencoders}

An autoencoder is a specialized neural network that learns to reconstruct input data through dimensional reduction. This reconstruction process involves compressing the input into a compact representation before expanding it back to its original form.

The network architecture comprises two main components: an encoder $E$ and a decoder $D$. The encoder maps input data to a lower-dimensional latent space $Z$, while the decoder reconstructs the original data from this compressed representation. The network can be trained using two approaches: direct reconstruction, where the objective is to minimize the error between input and output: $\theta = argmin_\theta |X - D(E(X))|$, or denoising reconstruction, where the network learns to recover original data from corrupted inputs $\hat{X}$: $\theta = argmin_\theta |X - D(E(\hat{X}))|$. Common corruption techniques include random masking of elements and the addition of Gaussian noise \cite{he2022masked}.

\begin{figure}[!htb]
    \centering
    \includegraphics[width=0.75\linewidth]{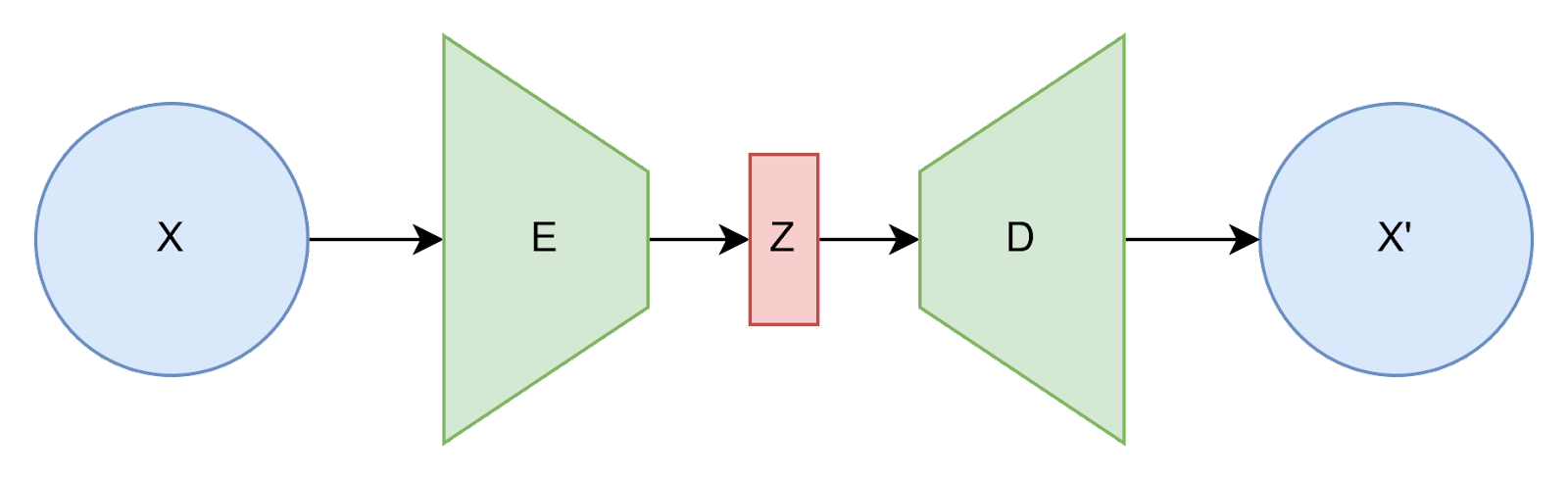}
    \caption{Autoencoder Architecture.}
    \label{fig:fig-ae}
\end{figure}

Following autoencoder training, the decoder $D$ functions as an independent model capable of sampling objects from space $X$ through transformations of objects from the low-dimensional latent space $Z$. The decoder $D$ operates as a generative model when the distribution of space $Z$ is known and sampling is feasible.

The classical autoencoder architecture, however, imposes no constraints on the latent vector space $Z$, rendering the sampling process impractical. Kingma and Welling addressed this limitation in their seminal work "Auto-Encoding Variational Bayes", introducing the variational autoencoder (VAE) framework \cite{kingma2013auto}.

The VAE operates by constraining the latent distribution Z to follow a standard normal distribution $z \sim \mathcal{N}(0;I)$. In this framework, the encoder and decoder learn transformations between given space and data space $X$: $E(x) = p(z|x), D(z)=p(z|x)$.

The authors demonstrate that the lower bound of maximum log-likelihood will be the following expression:
\begin{equation}
-L_{VAE} = \log p_{\theta}(x|z) - D_{KL}(p_\phi(z|x) || p(z)) \le \log p(x)
\end{equation}
In this equation $D_{KL}$ represents the Kullback-Leibler divergence, $p_{\theta}(x|z)$ denotes the decoder function $D$ that transforms latent space $Z$ to data space $X$, and $p_\phi(z|x)$ represents the encoder function $E$ that maps data space $X$ to latent space $Z$.

The VAE training process optimizes two objectives simultaneously: the reconstruction of input $x \in X$ and the alignment of the predicted latent distribution $Z'$ with the standard normal distribution $Z = \mathcal{N}(0;I)$.

The encoder generates a complete probability distribution, with the reconstruction scheme represented as $x' = D(z' \sim E(x))$. However, this approach presents a practical limitation since the sampling operation is non-differentiable, which prevents the implementation of the backpropagation algorithm.

To address this limitation, a constraint is applied to the encoder $E$ output, restricting it to predict parameters of the normal distribution -- specifically its mean and standard deviation. This enables sampling from the predicted distribution through the reparameterization trick, utilizing samples from the standard normal distribution: $x' = D(E_\sigma(x) \cdot (z'' \sim \mathcal{N}(0;I)) + E_\mu(x) )$. This formulation circumvents the non-differentiability issue during the optimization process.

\begin{figure} [!htb]
    \centering
    \includegraphics[width=0.75\linewidth]{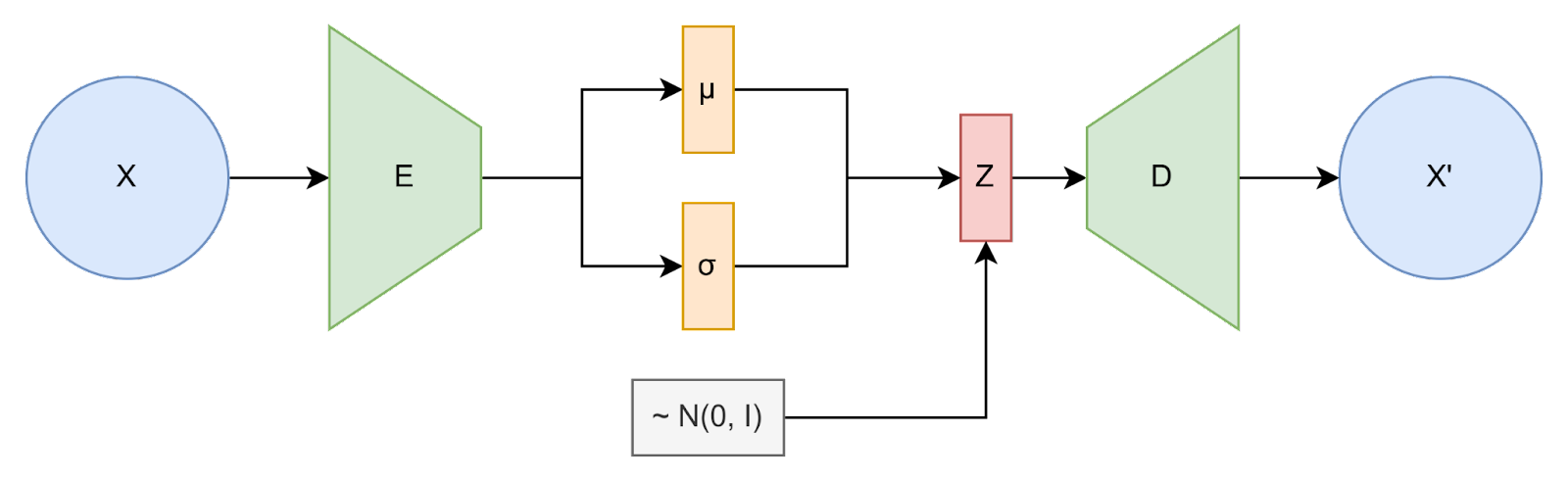}
    \caption{Variational autoencoder (VAE).}
    \label{fig:fig-vae}
\end{figure}

The discretization of latent space $Z$ presents an alternative approach to constraining autoencoder representations while preserving the decoder's sampling capabilities. Van den Oord et al. introduced this concept through the Vector Quantized Variational Autoencoder (VQ-VAE) in their work “Neural Discrete Representation Learning” \cite{vandenoord2017_repr}.

The proposed model architecture incorporates a discrete latent representation where each object $z \sim Z$ comprises vectors stored in an embedding codebook. This architecture ensures distribution matching between encoder predictions $E(x)$ and the prior distribution through the following quantization function:
\begin{equation}
z(x) = Quantize(E(x)) = e_k; k = argmin_j || E(x) - e_j ||
\end{equation}

However, two major technical challenges arise in this approach. First, the selection of elements from the quantized space is non-differentiable. Second, the determination of optimal fixed vectors within the quantized space remains problematic.

To address these challenges, the authors implement a dual-pulling mechanism. This mechanism simultaneously trains quantized space vectors by pulling them toward encoder $E$ predictions while drawing encoder predictions toward quantized values. This approach enables quantized vectors to approximate the most frequent non-quantized predictions of encoder $E$. The resulting loss function is formulated as:
\begin{equation}
L = || x - D(e_k) || + || sg[E(x)] - e_k || + \beta \cdot || E(x) - sg[e_k] ||,
\end{equation}
where $sg[\cdot]$ denotes the gradient stop operator during optimization.

The non-differentiability of quantization operations presents a notable challenge. However, this challenge is addressed by assuming $\nabla E(x) \approx \nabla e_k $, with the optimization function ensuring $E(x) \approx e_k$.

In recent years, variational autoencoders utilizing quantized vectors have proven particularly effective for semantic representation of complex objects. Their success has led to widespread adoption as decoders in latent space generative models \cite{rombach2021high}.

\subsection{Autoregressive models}

Sequential sampling presents an alternative approach to generating complete samples, whereby elements can be generated individually based on their predecessors. This method operates under the assumption that each element exhibits dependence solely on its antecedent elements. Consequently, the generative modeling task can be reformulated to approximate the conditional probability distribution:
\begin{equation}
    \prod_i p(x_i | x_1, x_2, ..., x_{i-1}) = \prod_i p(x_i | x_{<i}).
\end{equation}

Natural language modeling represents the predominant application of this sequential approach. Initially implemented in the foundational GPT architecture \cite{radford2018improving}, this methodology continues to serve as the cornerstone of contemporary "decoder-only" models.

Sequential modeling has emerged as a fundamental approach in natural language processing and showed its excellent efficiency with the introduction of the GPT architecture \cite{radford2018improving}. This methodology continues to be integral to contemporary "decoder-only" models \cite{achiam2023gpt, brown2020language, grattafiori2024llama}, which demonstrate remarkable capabilities in capturing long-range dependencies and generating contextually coherent outputs.

Research on scaling laws \cite{henighan2020scaling, hoffmann2022empirical, muennighoff2023scaling} has revealed systematic relationships between model performance and computational resources. These studies indicate that increases in model size and computational capacity correspond to improvements in cross-entropy loss across diverse domains. This pattern extends to image generation, video modeling, multimodal tasks, and mathematical problem-solving, following established power-law relationships \cite{xiong2025autoregressive}.

The demonstrated effectiveness of autoregressive models in natural language processing has facilitated their adoption in computer vision applications. Visual autoregressive models can be categorized into three distinct approaches based on their sequence representation mechanisms. The first approach, exemplified by Pixel-RNN \cite{vandenoord2016_pixelcnn}, implements pixel-by-pixel prediction through the transformation of 2D images into 1D pixel sequences. This method enables the modeling of both local and distant dependencies, albeit with significant computational requirements. The second approach utilizes token-based prediction, adapting techniques from natural language processing to employ discrete tokens for efficient high-resolution image processing, as demonstrated by VQ-GAN \cite{esser2020taming}. The third approach, represented by VAR \cite{tian2024visual}, implements scale-based prediction, generating content hierarchically across multiple resolutions in an autoregressive manner. Each of these approaches presents distinct advantages and limitations in practical applications \cite{xiong2025autoregressive}.

\subsection{Generative Adversarial Networks}

Generative models fundamentally rely on deriving optimization loss functions that characterize the likelihood of density distribution. However, Ian Goodfellow's seminal 2014 paper "Generative Adversarial Networks" introduced a novel paradigm: utilizing an additional neural network to learn the loss function \cite{goodfellow2014generative}.

The conventional approach of training a discriminative model for sample evaluation and subsequently using it to train the main model leads to rapid overfitting of the latter. To address this limitation, generative adversarial networks employ simultaneous training of both the main model and its discriminative counterpart, each with distinct loss functions. The primary model, responsible for generating target sample elements, is termed the generator, while the discriminative model that approximates the loss function is designated as the discriminator or critic.

\begin{figure} [!htb]
    \centering
    \includegraphics[width=0.75\linewidth]{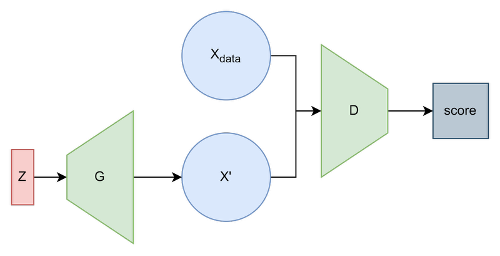}
    \caption{Conceptual framework of generative adversarial networks.}
    \label{fig:fig-gan}
\end{figure}

The discriminator's loss function is designed to predict a distinguishing score between generator-produced instances and those from the data distribution, thereby approximating the generator's loss function. The generator subsequently optimizes this same score with the objective of making its generated instances indistinguishable from the data distribution instances. Through this process, the generator progressively aligns its generative distribution with the target data distribution.

Goodfellow's seminal work conceptualizes the discriminator as a binary classifier, employing binary cross-entropy as the loss function -- specifically, the negative log-likelihood of the binomial distribution \cite{goodfellow2014generative}. This adversarial training framework can be formally expressed as a minimax game:
\begin{equation}
    \min_G \max_D (E_{x \sim p_{data}}[\log D(x)] + E_{z \sim p_Z}[1-\log D(G(z))]).
\end{equation}

The practical implementation involves alternating optimization steps between the discriminator and generator components. A notable optimization is that the first term, being independent of the generator, can be omitted during the generator's optimization phase.

Goodfellow identifies a potential limitation of this minimax game structure. The issue arises when the discriminator exhibits high confidence in identifying generated instances as synthetic. Under these conditions, the derivative of the second term approaches zero, effectively preventing the generator from receiving meaningful training signals. To address this limitation, Goodfellow proposes a "non-saturating adversarial loss function" \cite{goodfellow2014generative}, defined as:
\begin{equation}
    \max_D (E_{x \sim p_{data}}[\log D(x)] + E_{z \sim p_Z} [1-\log D(G(z))])
\end{equation}
\begin{equation}
    \min_G (- E_{z \sim p_Z} [\log D(G(z))])
\end{equation}

Generative Adversarial Networks (GANs) have emerged as a versatile tool across multiple domains in contemporary research. In the field of computer vision, GANs demonstrate particular utility in high-quality image generation, resolution enhancement, and style transfer applications. These capabilities facilitate the advancement of automated content creation systems, medical image enhancement for diagnostic purposes, and the development of photorealistic virtual environments. In the industrial sector, GANs have proven instrumental in synthetic data generation, enabling effective machine learning model training while circumventing the need for extensive real-world data collection \cite{abou2024generative, sharma2024generative}.

Recent advances in GAN research primarily address architectural improvements and loss function optimization to enhance training stability and efficiency. The introduction of advanced variants, such as StyleGAN3, has yielded significant improvements in the quality and controllability of generated images and videos. Furthermore, the application scope of GANs extends to cybersecurity applications, specifically in malware detection and prevention, as well as financial market modeling and anomaly detection systems \cite{chakraborty2024ten, dunmore2023generative}.

\subsection{Normalizing flows}

Normalizing flows constitute a family of generative models that facilitate both efficient sampling and accurate density estimation. These models have proven effective in diverse applications: image synthesis, noise modeling, video and audio generation, graph construction, reinforcement learning, computer graphics, and physical systems simulation \cite{kobyzev2020normalizing}.

The essence of normalizing flows is the transformation of a simple distribution through a sequence of invertible and differentiable functions. The invertibility property enables the estimation of probability density for samples drawn from the target distribution. This estimation involves transforming the sample into a simple distribution and computing the product of two terms: the density of the inversely transformed sample and the volume change induced by the sequence of inverse transformations \cite{rezende2015variational}.

This relationship can be formally expressed as:
\begin{equation}
    p(x) = p(f^{-1}(x))| det(\frac{\partial f^{-1}(x)}{\partial x}) | = p(z_k) | det(\frac{\partial z_k}{\partial x}) |.  
\end{equation}

The transformation can be extended to the entire sequence of functions, yielding a direct likelihood estimate:
\begin{equation}
    \log p(x) = \log p(z_0) - \sum_i \log | det (\frac{\partial f^{-1}_{i+1}(z_i)}{\partial z_i}) |
\end{equation}

This formulation enables the optimization of normalizing flows through direct maximization of the input data likelihood.

\begin{figure} [!htb]
    \centering
    \includegraphics[width=0.75\linewidth]{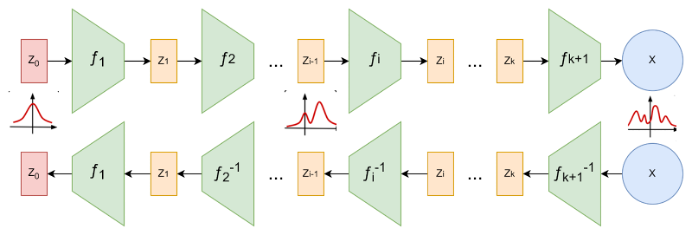}
    \caption{Conceptual framework of normalizing flows.}
    \label{fig:fig-normflow}
\end{figure}

The implementation of functions $f_i$ through complex neural networks enables sufficient approximation of complex transformations. However, practical applications require these functions to possess two critical properties: invertibility and efficient Jacobian computation. These requirements impose significant architectural constraints on the applicable neural network structures.

The affine coupling layer serves as a notable example of such functions, representing an invertible bijective transformation. This function implements a strategic division of the input vector into two components: the first component remains unaltered, while the second undergoes an affine transformation parameterized by the first component \cite{dinh2016density}.

The transformation can be expressed as:
\begin{equation}
    y_{1:d} = x_{1:d}, y_{d+1:n} = x_{d+1:n} * e^{scale(x_{1:d})} + t(x_{1:d}).
\end{equation}

The inverse transformation is defined as:
\begin{equation}
    x_{1:d} = y_{1:d}, x_{d+1:n} = (y_{d+1:n} - t(x_{1:d})) * e^{-scale(x_{1:d})}.
\end{equation}

A distinctive characteristic of this function lies in its Jacobian structure, where the determinant is computed as the product of diagonal elements, with the first d diagonal entries being ones.

\begin{equation}
    J = \begin{bmatrix} I_d & 0_{d \times (n-d) } \\ \frac{\partial y_{d+1:n}}{\partial x_{1:d}} & diag(e^{s(x_{1:d})}) \end{bmatrix}
\end{equation}

The development and optimization of such function implementations remain an active research area, as evidenced by numerous scientific publications \cite{kobyzev2020normalizing}.

\subsection{Diffusion models and flow matching}

The concept of diffusion models emerged through parallel developments under various names and theoretical frameworks: diffusion probabilistic models \cite{pmlr-v37-sohl-dickstein15}, noise conditional score networks \cite{Song2019_diffusion}, and denoising diffusion probabilistic models \cite{ho2020denoising}. Despite their different origins, these approaches converge to a unified theoretical framework now known as diffusion models.

Diffusion models have gained significant prominence in contemporary machine learning applications, demonstrating versatility across multiple domains including image synthesis, video generation, audio processing, motion prediction, and animation rendering. The field experienced a pivotal advancement following the publication of Dhariwal and Nichol's seminal work "Diffusion Models Beat GANs on Image Synthesis", which empirically demonstrated the superior performance of diffusion models compared to generative adversarial networks in image synthesis tasks \cite{dhariwal2021diffusion}.

Diffusion models represent a class of generative models that transform complex data distributions into a standard Gaussian distribution through an iterative process of noise injection and rescaling. This forward diffusion process can be formally expressed as:
\begin{equation}
    x_t = scale(t) \cdot x_{t-1} + n(t) \cdot \mathcal{N}(0;I)
\end{equation}
where successive applications of this transformation gradually convert the initial distribution into a standard normal distribution. The core mechanism of diffusion models lies in their ability to learn and approximate the reverse diffusion process, which aims to systematically extract meaningful signals from noise-corrupted data.

\begin{figure} [!tbh]
    \centering
    \includegraphics[width=0.75\linewidth]{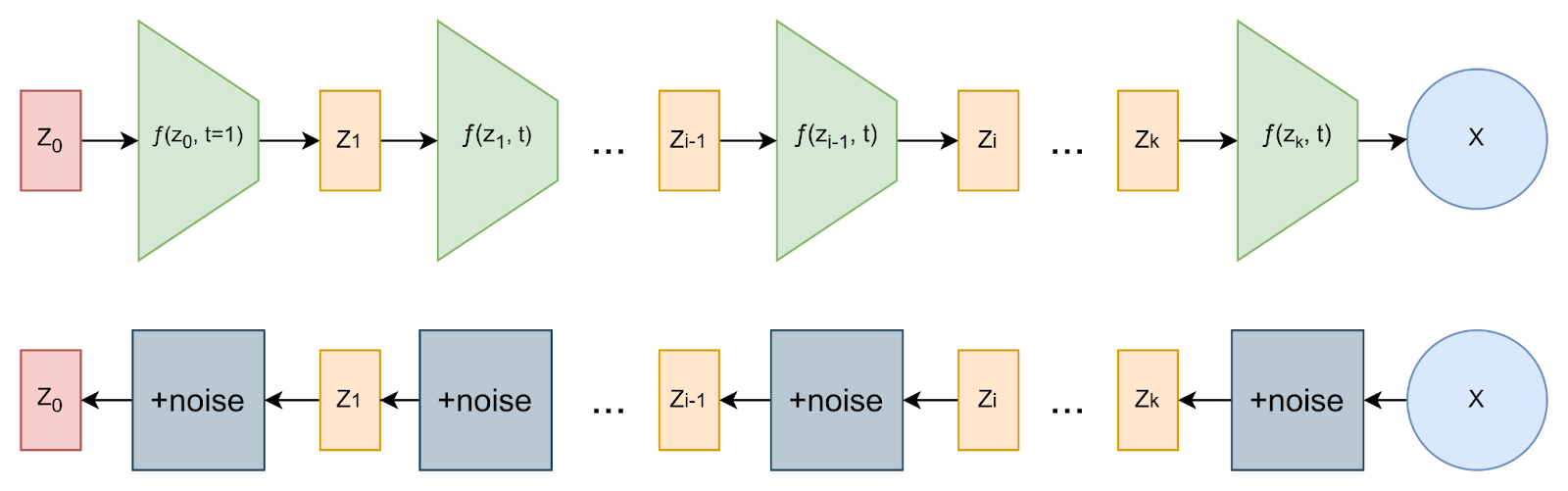}
    \caption{Conceptual framework of diffusion models.}
    \label{fig:fig-diff}
\end{figure}

The forward transformation formulation enables direct noisy data sampling at any step through a closed-form expression, leveraging the property that the sum of normal distributions yields a normal distribution. The reverse transformation process involves iterative applications of a single function that processes a noisy sample and a time parameter $t$, which indicates the sample's noise level. The conventional approach assigns $t=0$ to noiseless samples and $t=1$ to completely noisy states where the original signal is absent.

The diffusion model operates through a time-dependent vector transformation function, which can be expressed mathematically as $\Delta x = f_\theta (x) \Delta t$ or $f_\theta=\frac{dx}{dt}$. This formulation indicates that the transformation function represents a solution to a specific
differential equation.

The differential equation framework of the target function enables the application of established mathematical approaches in this domain \cite{karras2022elucidating}. Song and Ermonth conceptualize the diffusion model as a score function approximating the gradient of the log-probability density $\nabla \log p_{data}(x)$. Their methodology primarily relies on two fundamental components: score estimation and Langevin dynamics \cite{Song2019_diffusion}.

The optimization problem requires minimizing $E _{p_{data}} | f_\theta (x) - \nabla_x \log p_{data}(x) |$, which transforms into the minimization of $E _{p_{data}} [tr(\nabla_x f_\theta(x)) + \frac{1}{2}|f_\theta (x)|]$. However, this formulation presents practical limitations due to the computational complexity of calculating the Jacobian trace for our sophisticated function. Song and Ermonth address this challenge by introducing a noise distribution $q_\sigma(\hat{x}|x)$ and employing approximation techniques, resulting in a more tractable loss function  $E _{p_{data}} | f_\theta (x) - \nabla_{\hat{x}} \log q_\sigma(\hat{x}|x) |$ \cite{Song2019_diffusion}.

Langevin dynamics enables sampling from a probability distribution $p(x)$ through the score function $\nabla_x \log p(x)$ and the iterative update rule:
\begin{equation}
    \hat{x}_t = \hat{x}_{t-1} + 2\nabla_x \log p(\hat{x}_{t-1}) + \sqrt{\epsilon} z_t, \text{where }
    z_t \sim \mathcal{N}(0; I).
\end{equation}

The convergence of distribution $\hat{x}_t$ to $p(x)$ occurs under the conditions $\epsilon \xrightarrow{} 0, T \xrightarrow{} \infty$. By substituting our neural network as the score function, we can obtain a generative model that samples from the data distribution \cite{Song2019_diffusion}.

At its core, Langevin dynamics serves as a numerical solution method for stochastic differential equations. Research by Karras et al. establishes the independence of the sampling process from score function optimization. This independence permits the application of various numerical solvers, encompassing both stochastic differential equations (SDE) and ordinary differential equations (ODE), under the assumption of zero Wiener noise in Langevin dynamics \cite{karras2022elucidating}.

Flow matching represents an alternative formulation of diffusion models. This approach characterizes the generation process through continuous normalizing flows, incorporating a linearity constraint on the velocity vector field that facilitates bidirectional movement between distribution objects \cite{lipman2022flow}. The resulting iterative process resembles the diffusion process and can be expressed as:
\begin{equation}
    x_t = t x_0 + (1-t)y.
\end{equation}

In this equation, $x_0$ represents an element of the target distribution and $y$ denotes the corresponding element of the source distribution Y. While the source distribution $Y$ can be normal, flow matching models offer greater flexibility in distribution selection compared to traditional diffusion models.

The training methodology for flow matching models relies on a known vector field that is subsequently learned through neural network optimization. Researchers demonstrated that the expected velocity of movement between random elements across distributions equals the expected velocity between corresponding elements from the same distributions. This property enables the computation of the loss function using random distribution pairs:
\begin{equation}
    L = E_{t,x \sim X,y \sim Y} (f_\theta(x_t)-u_t) = E_{t,x \sim X,y \sim Y} (f_\theta(t x_0 + (1-t)y) - (x-y)).
\end{equation}

However, flow matching models exhibit longer training time compared to diffusion models. Recent research suggests that this limitation might be addressed through optimal pair selection methodologies, particularly through optimal transport techniques \cite{shi2023diffusion, tong2024improving}.

\section{Generative models as the probability transformation functions}

The probability transformation function constitutes a mathematical operation that maps a random variable from its initial probability distribution to a new variable with a target distribution. This transformation represents a fundamental concept in probability theory, enabling the modification of probability distributions -- a crucial capability for modeling complex real-world phenomena. In formal mathematical terms, given a random variable $X$ with a probability density function $p(x)$ and a transformation function $f$ that yields a new variable $y=f(x)$, the probability density of the transformed variable $p_Y(y)$ is expressed as:
\begin{equation}
    p_Y(y) = p_X(f^{-1}(y)) \lvert det( \frac{\partial}{\partial y}f^{-1}(y) ) \rvert.
\end{equation}

The change of variables formula maintains the conservation of probability mass throughout the transformation process, ensuring that the integral of the density function equals unity \cite{chen2023new}.

We claim that this mathematical framework of probability transformation functions provides the theoretical foundation for generative models, which learn to map simple probability distributions to more complex ones.

\textbf{Normalizing flows} represent the most direct implementation of the distribution transformation function concept. The fundamental principle of this generative model lies in learning a function $F$ that maps the distribution of observed data to a normal distribution, expressed as $F(y \sim D) = x$. This mapping is constrained by the requirement of invertibility, enabling the generation of complex distribution elements through the inverse transformation $F^{-1}(x \sim D)$, where elements from a normal distribution serve as input. Furthermore, the training methodology for normalizing flows is theoretically grounded in the principle of distribution mass conservation during the transformation process.

We now turn our attention to \textbf{diffusion models} and \textbf{flow matching} models. Flow matching represents a generalization of diffusion models while operating within the framework of continuous normalizing flows \cite{lipman2022flow}. These approaches are fundamentally based on vector fields that define differential equations describing the transformation path between distributions.

The core mechanism of these models involves studying vector fields that govern the transformation dynamics at each point in data space and time. While the training and optimization methodologies vary, the general sampling process follows a consistent pattern: an element is sampled from an initial distribution (Gaussian for diffusion models), and undergoes sequential transformations. These transformations are parameterized by their position along the temporal path, ultimately converging to an element in the target distribution.

This process can be formally expressed as a composite function $y=F(x), x \sim D_0$,
where each $f_{t_i}$ represents a transformation at time step $t_i, i=\underline{0,n}$. This formulation shares fundamental similarities with normalizing flows, with a notable distinction: the approach focuses on partial aspects of the composite function and allows for variable iteration counts based on specific requirements. Consequently, the complete function $F(x \sim D_0)$ serves as a distribution transformation function, mapping elements from the source distribution to the target distribution.

\textbf{Generative Adversarial Networks} represent a distinct approach compared to the previously examined models. The fundamental concept of GANs relies on a minimax game between two neural networks, which deviates from conventional probability distribution transformations at first glance. But during the inference, the process exclusively utilizes the generator $G$, which accepts an input element from a predefined distribution, typically Gaussian normal, and produces an element from the target distribution, expressed as $y = G(x \sim N)$. Thus, despite the difference in training, the GAN generator can be viewed as a probability distribution transformation function during inference, similar to other generative models.

\textbf{Autoencoders}, including their variational and quantized variants, can be analyzed through a similar framework. The inference phase includes only the decoder component, which receives input from a predetermined distribution, such as normal or quantized distributions. The decoder functions as a distribution transformation mechanism, represented as $y = D(z \sim Z)$, converting the known input distribution into the target distribution.

While \textbf{autoregressive models} lack an explicit latent distribution as the generation process foundation, their implementation reveals a more complex mechanism. A detailed examination of these models demonstrates that the prediction of subsequent elements incorporates both historical elements and stochastic components. The stochastic elements are essential for ensuring prediction diversity in autoregressive models. Contemporary approaches, including beam search, top-k sampling, and nucleus sampling \cite{wiher-etal-2022-decoding}, predominantly employ random variables drawn from normal or uniform distributions for token selection.

This generation process can be formally expressed as
\begin{equation}
    y_n = f(y_{n-1},y_{n-2},\dots,y_1,z \sim U) = f'_n(z), z \sim U(0,1),
\end{equation}
where $z$ represents the random variable used for sampling. When expanding the predictions for all preceding elements, the equation becomes

\begin{equation}
\begin{split}
    y_n = f'_n(z_n) \circ f'_{n-1}(z_{n-1}) \dots \circ f'_1(x_1, z_1) = F(x_1,z_1,z_2,\dots,z_n), \\
    y_n = F(x_1, (z_n)), \text{where } (z_n) \sim U(0,1).
\end{split}
\end{equation}

This formulation reveals that autoregressive models inherently function as distribution transformation mechanisms, accepting sampling distributions as inputs and transforming them into elements from the target distribution.

A crucial engineering consideration underlies the widespread implementation of latent spaces with predetermined distributions or stochastic element selection in generative models. The incorporation of stochastic elements is essential for these models to generate diverse outputs, as purely deterministic calculations would invariably produce identical results. This research presents an alternative perspective on this universal characteristic of generative models by examining them as members of the distribution transformation function class.

\begin{figure} [!tbh]
    \centering
    \includegraphics[width=0.75\linewidth]{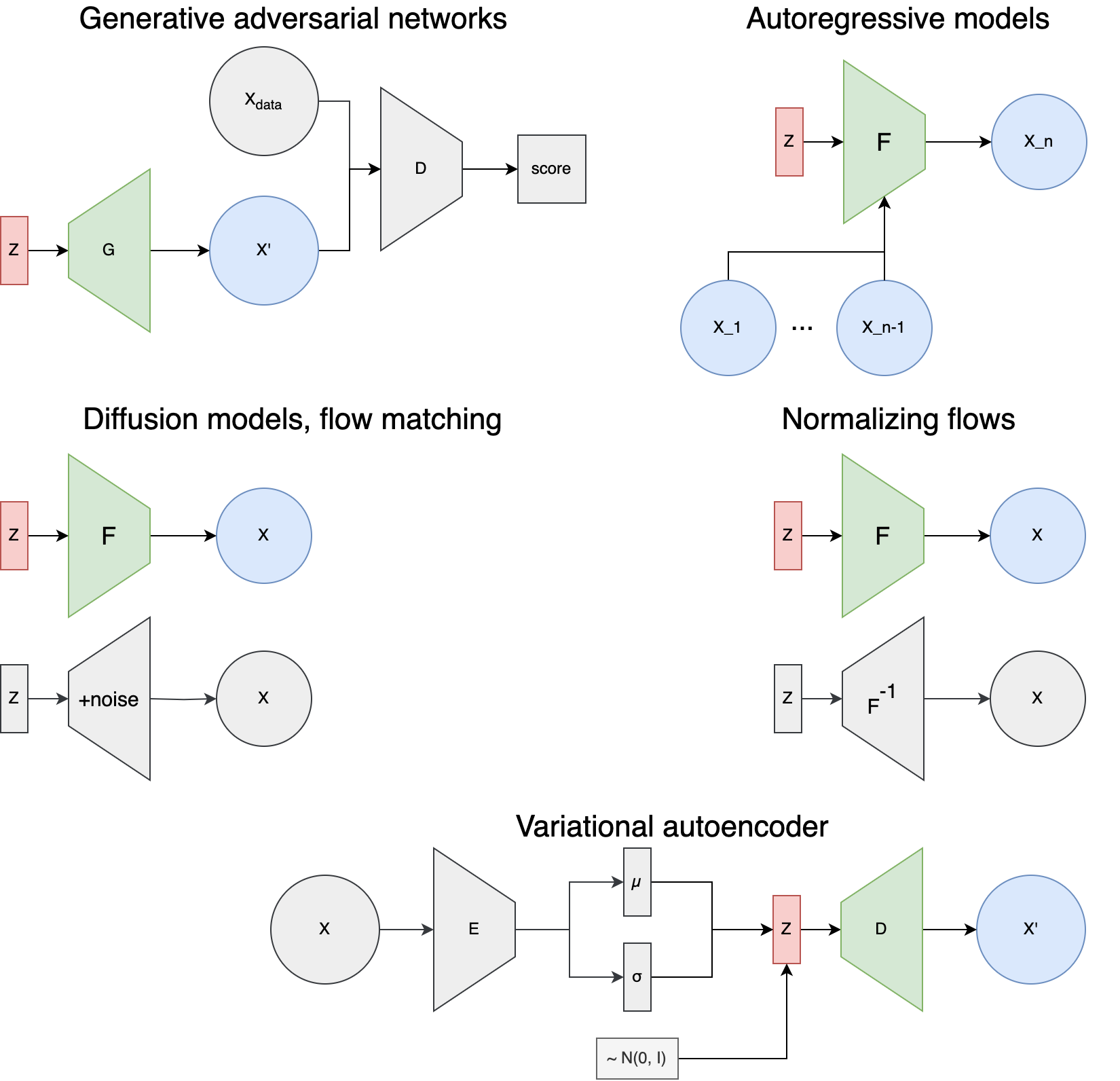}
    \caption{Visualization of different generative model types with color-coded elements aligned for comparison.}
    \label{fig:fig-all}
\end{figure}

Figure \ref{fig:fig-all} presents a novel visualization of the primary generative models, featuring an unconventional arrangement where latent and target spaces are aligned across different models. The non-inference elements are distinguished through gray shading. For visualization purposes, the multi-step transformations in normalizing flows and diffusion models are represented by a single generalized function.

This aligned representation emphasizes a fundamental concept: the core component of numerous generative models is a probability transformation function that maps a latent distribution to a target data distribution. The latent space invariably consists of a predefined simple distribution. The distinguishing features among these models primarily serve as auxiliary mechanisms, each implementing different approaches to establish loss functions for optimizing the generative transformation.

\section{Conclusion}

In this paper, we have presented a unified theoretical perspective on deep generative models by demonstrating that they fundamentally operate as probability transformation functions. Through careful examination of various architectures -- from autoencoders and GANs to diffusion models and normalizing flows -- we have shown that despite their apparent differences in training methodologies and architectures, these models share a common underlying principle: they transform simple predefined distributions into complex target data distributions.

This unifying framework provides valuable insights into the fundamental nature of generative models. While these models have traditionally been studied as separate domains with distinct theoretical foundations, our analysis reveals that their core operational principle remains consistent across different architectures. The distinguishing features among these models primarily serve as auxiliary mechanisms for optimizing the transformation function, rather than representing fundamentally different approaches to generation.

This perspective opens several promising directions for future research. First, it suggests the possibility of developing new theoretical approaches that can be applied universally across different types of generative models. Additionally, this unified view may facilitate the transfer of methodological improvements between different model architectures, potentially leading to more efficient and effective generative modeling techniques. Further investigation of the mathematical properties of these transformation functions could also yield insights into the theoretical capabilities and limitations of different generative approaches.

\bibliographystyle{unsrt}  
\bibliography{references}

\begin{thebibliography}{10}

\bibitem{bond2021deep}
Sam Bond-Taylor, Adam Leach, Yang Long, and Chris~G Willcocks.
\newblock Deep generative modelling: A comparative review of vaes, gans, normalizing flows, energy-based and autoregressive models.
\newblock {\em IEEE transactions on pattern analysis and machine intelligence}, 44(11):7327--7347, 2021.

\bibitem{yang2024_diff}
Ling Yang, Zhilong Zhang, Yang Song, Shenda Hong, Runsheng Xu, Yue Zhao, Wentao Zhang, Bin Cui, and {Ming Hsuan} Yang.
\newblock Diffusion models: A comprehensive survey of methods and applications.
\newblock {\em ACM Computing Surveys}, 56(4), April 2024.
\newblock Publisher Copyright: {\textcopyright} 2023 held by the owner/author(s). Publication rights licensed to ACM.

\bibitem{gozalo2023survey}
Roberto Gozalo-Brizuela and Eduardo~C Garrido-Merch{\'a}n.
\newblock A survey of generative ai applications.
\newblock {\em arXiv preprint arXiv:2306.02781}, 2023.

\bibitem{lipman2022flow}
Yaron Lipman, Ricky~TQ Chen, Heli Ben-Hamu, Maximilian Nickel, and Matt Le.
\newblock Flow matching for generative modeling.
\newblock {\em arXiv preprint arXiv:2210.02747}, 2022.

\bibitem{goodfellow2016nips}
Ian Goodfellow.
\newblock Nips 2016 tutorial: Generative adversarial networks.
\newblock {\em arXiv preprint arXiv:1701.00160}, 2016.

\bibitem{he2022masked}
Kaiming He, Xinlei Chen, Saining Xie, Yanghao Li, Piotr Doll{\'a}r, and Ross Girshick.
\newblock Masked autoencoders are scalable vision learners.
\newblock In {\em Proceedings of the IEEE/CVF conference on computer vision and pattern recognition}, pages 16000--16009, 2022.

\bibitem{kingma2013auto}
Diederik~P Kingma, Max Welling, et~al.
\newblock Auto-encoding variational bayes, 2013.

\bibitem{vandenoord2017_repr}
Aaron van~den Oord, Oriol Vinyals, and koray kavukcuoglu.
\newblock Neural discrete representation learning.
\newblock In I.~Guyon, U.~Von Luxburg, S.~Bengio, H.~Wallach, R.~Fergus, S.~Vishwanathan, and R.~Garnett, editors, {\em Advances in Neural Information Processing Systems}, volume~30. Curran Associates, Inc., 2017.

\bibitem{rombach2021high}
Robin Rombach, Andreas Blattmann, Dominik Lorenz, Patrick Esser, and Bj{\"o}rn Ommer.
\newblock High-resolution image synthesis with latent diffusion models, 2021, 2021.

\bibitem{radford2018improving}
Alec Radford, Karthik Narasimhan, Tim Salimans, Ilya Sutskever, et~al.
\newblock Improving language understanding by generative pre-training.
\newblock 2018.

\bibitem{achiam2023gpt}
Josh Achiam, Steven Adler, Sandhini Agarwal, Lama Ahmad, Ilge Akkaya, Florencia~Leoni Aleman, Diogo Almeida, Janko Altenschmidt, Sam Altman, Shyamal Anadkat, et~al.
\newblock Gpt-4 technical report.
\newblock {\em arXiv preprint arXiv:2303.08774}, 2023.

\bibitem{brown2020language}
Tom Brown, Benjamin Mann, Nick Ryder, Melanie Subbiah, Jared~D Kaplan, Prafulla Dhariwal, Arvind Neelakantan, Pranav Shyam, Girish Sastry, Amanda Askell, et~al.
\newblock Language models are few-shot learners.
\newblock {\em Advances in neural information processing systems}, 33:1877--1901, 2020.

\bibitem{grattafiori2024llama}
Aaron Grattafiori, Abhimanyu Dubey, Abhinav Jauhri, Abhinav Pandey, Abhishek Kadian, Ahmad Al-Dahle, Aiesha Letman, Akhil Mathur, Alan Schelten, Alex Vaughan, et~al.
\newblock The llama 3 herd of models.
\newblock {\em arXiv preprint arXiv:2407.21783}, 2024.

\bibitem{henighan2020scaling}
Tom Henighan, Jared Kaplan, Mor Katz, Mark Chen, Christopher Hesse, Jacob Jackson, Heewoo Jun, Tom~B Brown, Prafulla Dhariwal, Scott Gray, et~al.
\newblock Scaling laws for autoregressive generative modeling.
\newblock {\em arXiv preprint arXiv:2010.14701}, 2020.

\bibitem{hoffmann2022empirical}
Jordan Hoffmann, Sebastian Borgeaud, Arthur Mensch, Elena Buchatskaya, Trevor Cai, Eliza Rutherford, Diego de~Las~Casas, Lisa~Anne Hendricks, Johannes Welbl, Aidan Clark, et~al.
\newblock An empirical analysis of compute-optimal large language model training.
\newblock {\em Advances in neural information processing systems}, 35:30016--30030, 2022.

\bibitem{muennighoff2023scaling}
Niklas Muennighoff, Alexander Rush, Boaz Barak, Teven Le~Scao, Nouamane Tazi, Aleksandra Piktus, Sampo Pyysalo, Thomas Wolf, and Colin~A Raffel.
\newblock Scaling data-constrained language models.
\newblock {\em Advances in Neural Information Processing Systems}, 36:50358--50376, 2023.

\bibitem{xiong2025autoregressive}
Jing Xiong, Gongye Liu, Lun Huang, Chengyue Wu, Taiqiang Wu, Yao Mu, Yuan Yao, Hui Shen, Zhongwei Wan, Jinfa Huang, Chaofan Tao, Shen Yan, Huaxiu Yao, Lingpeng Kong, Hongxia Yang, Mi~Zhang, Guillermo Sapiro, Jiebo Luo, Ping Luo, and Ngai Wong.
\newblock Autoregressive models in vision: A survey.
\newblock {\em Transactions on Machine Learning Research}, 2025.
\newblock Survey Certification.

\bibitem{vandenoord2016_pixelcnn}
Aaron van~den Oord, Nal Kalchbrenner, Lasse Espeholt, koray kavukcuoglu, Oriol Vinyals, and Alex Graves.
\newblock Conditional image generation with pixelcnn decoders.
\newblock In D.~Lee, M.~Sugiyama, U.~Luxburg, I.~Guyon, and R.~Garnett, editors, {\em Advances in Neural Information Processing Systems}, volume~29. Curran Associates, Inc., 2016.

\bibitem{esser2020taming}
Patrick Esser, Robin Rombach, and Bj{\"o}rn Ommer.
\newblock Taming transformers for high-resolution image synthesis. 2021 ieee.
\newblock In {\em CVF Conference on Computer Vision and Pattern Recognition (CVPR)}, volume~10, 2020.

\bibitem{tian2024visual}
Keyu Tian, Yi~Jiang, Zehuan Yuan, BINGYUE PENG, and Liwei Wang.
\newblock Visual autoregressive modeling: Scalable image generation via next-scale prediction.
\newblock In {\em The Thirty-eighth Annual Conference on Neural Information Processing Systems}, 2024.

\bibitem{goodfellow2014generative}
Ian~J Goodfellow, Jean Pouget-Abadie, Mehdi Mirza, Bing Xu, David Warde-Farley, Sherjil Ozair, Aaron Courville, and Yoshua Bengio.
\newblock Generative adversarial nets.
\newblock {\em Advances in neural information processing systems}, 27, 2014.

\bibitem{abou2024generative}
Chafic Abou~Akar, Rachelle Abdel~Massih, Anthony Yaghi, Joe Khalil, Marc Kamradt, and Abdallah Makhoul.
\newblock Generative adversarial network applications in industry 4.0: A review.
\newblock {\em International Journal of Computer Vision}, 132(6):2195--2254, 2024.

\bibitem{sharma2024generative}
Preeti Sharma, Manoj Kumar, Hitesh~Kumar Sharma, and Soly~Mathew Biju.
\newblock Generative adversarial networks (gans): introduction, taxonomy, variants, limitations, and applications.
\newblock {\em Multimedia Tools and Applications}, pages 1--48, 2024.

\bibitem{chakraborty2024ten}
Tanujit Chakraborty, Ujjwal~Reddy KS, Shraddha~M Naik, Madhurima Panja, and Bayapureddy Manvitha.
\newblock Ten years of generative adversarial nets (gans): a survey of the state-of-the-art.
\newblock {\em Machine Learning: Science and Technology}, 5(1):011001, 2024.

\bibitem{dunmore2023generative}
Aeryn Dunmore, Julian Jang-Jaccard, Fariza Sabrina, and Jin Kwak.
\newblock Generative adversarial networks for malware detection: a survey.
\newblock {\em arXiv preprint arXiv:2302.08558}, 2023.

\bibitem{kobyzev2020normalizing}
Ivan Kobyzev, Simon~JD Prince, and Marcus~A Brubaker.
\newblock Normalizing flows: An introduction and review of current methods.
\newblock {\em IEEE transactions on pattern analysis and machine intelligence}, 43(11):3964--3979, 2020.

\bibitem{rezende2015variational}
Danilo Rezende and Shakir Mohamed.
\newblock Variational inference with normalizing flows.
\newblock In {\em International conference on machine learning}, pages 1530--1538. PMLR, 2015.

\bibitem{dinh2016density}
Laurent Dinh, Jascha Sohl-Dickstein, and Samy Bengio.
\newblock Density estimation using real nvp.
\newblock {\em arXiv preprint arXiv:1605.08803}, 2016.

\bibitem{pmlr-v37-sohl-dickstein15}
Jascha Sohl-Dickstein, Eric Weiss, Niru Maheswaranathan, and Surya Ganguli.
\newblock Deep unsupervised learning using nonequilibrium thermodynamics.
\newblock In Francis Bach and David Blei, editors, {\em Proceedings of the 32nd International Conference on Machine Learning}, volume~37 of {\em Proceedings of Machine Learning Research}, pages 2256--2265, Lille, France, 07--09 Jul 2015. PMLR.

\bibitem{Song2019_diffusion}
Yang Song and Stefano Ermon.
\newblock Generative modeling by estimating gradients of the data distribution.
\newblock In H.~Wallach, H.~Larochelle, A.~Beygelzimer, F.~d\textquotesingle Alch\'{e}-Buc, E.~Fox, and R.~Garnett, editors, {\em Advances in Neural Information Processing Systems}, volume~32. Curran Associates, Inc., 2019.

\bibitem{ho2020denoising}
Jonathan Ho, Ajay Jain, and Pieter Abbeel.
\newblock Denoising diffusion probabilistic models.
\newblock {\em Advances in neural information processing systems}, 33:6840--6851, 2020.

\bibitem{dhariwal2021diffusion}
Prafulla Dhariwal and Alexander Nichol.
\newblock Diffusion models beat gans on image synthesis.
\newblock {\em Advances in neural information processing systems}, 34:8780--8794, 2021.

\bibitem{karras2022elucidating}
Tero Karras, Miika Aittala, Timo Aila, and Samuli Laine.
\newblock Elucidating the design space of diffusion-based generative models.
\newblock {\em Advances in neural information processing systems}, 35:26565--26577, 2022.

\bibitem{shi2023diffusion}
Yuyang Shi, Valentin De~Bortoli, Andrew Campbell, and Arnaud Doucet.
\newblock Diffusion schr{\"o}dinger bridge matching.
\newblock {\em Advances in Neural Information Processing Systems}, 36:62183--62223, 2023.

\bibitem{tong2024improving}
Alexander Tong, Kilian FATRAS, Nikolay Malkin, Guillaume Huguet, Yanlei Zhang, Jarrid Rector-Brooks, Guy Wolf, and Yoshua Bengio.
\newblock Improving and generalizing flow-based generative models with minibatch optimal transport.
\newblock {\em Transactions on Machine Learning Research}, 2024.
\newblock Expert Certification.

\bibitem{chen2023new}
Luyuan Chen and Yong Deng.
\newblock A new probability transformation approach of mass function.
\newblock {\em Soft Computing}, 27(20):15123--15132, 2023.

\bibitem{wiher-etal-2022-decoding}
Gian Wiher, Clara Meister, and Ryan Cotterell.
\newblock On decoding strategies for neural text generators.
\newblock {\em Transactions of the Association for Computational Linguistics}, 10:997--1012, 2022.

\end{thebibliography}

\end{document}